\documentclass[10pt,twocolumn,letterpaper]{article}

\usepackage{cvpr}              

\usepackage{cuted}
\usepackage{stfloats}
\usepackage{placeins}
\usepackage{multirow}

\newcommand{\shortcite}{\cite}

\newcommand{\revise}[1]{{#1}}

\definecolor{cvprblue}{rgb}{0.21,0.49,0.74}
\usepackage[pagebackref,breaklinks,colorlinks,allcolors=cvprblue]{hyperref}

\def\paperID{9517} 
\def\confName{CVPR}
\def\confYear{2026}

\title{FabricGen: Microstructure-Aware Woven Fabric Generation}

\author{
    Yingjie Tang$^{1\dagger}$ \quad 
    Di Luo$^{1\dagger}$ \quad 
    Zixiong Wang$^{1}$ \quad 
    Xiaoli Ling$^{2}$ \quad 
    Jian Yang$^{1,2}$ \quad 
    Beibei Wang$^{2*}$ \\
    {\normalsize $^{1}$Nankai University} \quad {\normalsize $^{2}$Nanjing University}
}

\usepackage{mathtools}
\usepackage{xspace}

\begin{document}
\maketitle

\begin{strip}
    \centering
    \vspace{-40px}
    \includegraphics[width=1.0\textwidth]{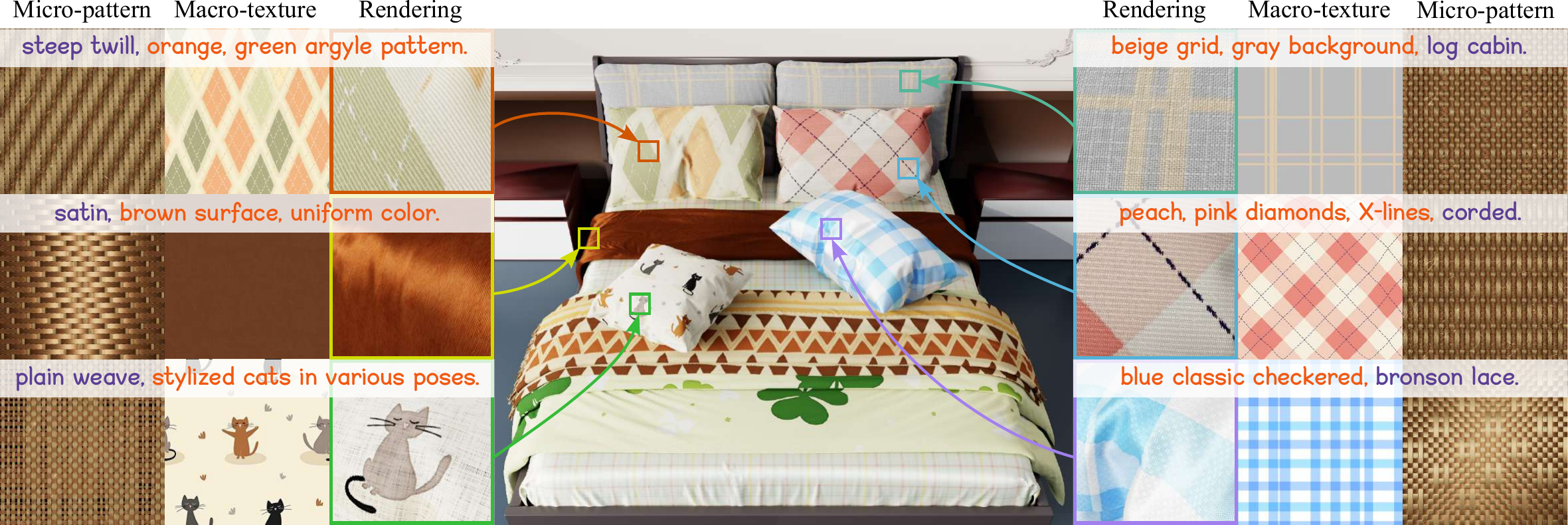}
    \captionof{figure}{
We present FabricGen, an end-to-end \revise{material generation} framework that generates woven fabric materials with fine details, enabling non-expert users to create diverse and highly detailed fabric materials for photorealistic rendering. 
    }
\label{fig:teaser}
\end{strip}

\newcommand\blfootnote[1]{%
  \begingroup
  \renewcommand\thefootnote{}\footnote{#1}%
  \addtocounter{footnote}{-1}%
  \endgroup
}
\blfootnote{* Corresponding author.}
\blfootnote{$\dagger$ These authors contribute equally.}
\blfootnote{$^1$ College of Computer Science, Nankai University, Tianjin, China}
\blfootnote{$^2$ School of Intelligence Science and Technology, Nanjing University, Suzhou, China}

\begin{abstract}
Woven fabric materials are widely used in rendering applications, yet designing realistic examples typically involves multiple stages, requiring expertise in weaving principles and texture authoring. Recent advances have explored diffusion models to streamline this process. However, pre-trained diffusion models often struggle to generate intricate yarn-level details that conform to weaving rules. To address this, we present \emph{FabricGen}, an end-to-end framework for generating high-quality woven fabric materials from textual descriptions. A key insight of our method is the decomposition of macro-scale textures and micro-scale weaving patterns. To generate macro-scale textures free from microstructures, we fine-tune pre-trained diffusion models on a collected dataset of microstructure-free fabrics. As for micro-scale weaving patterns, we develop an enhanced procedural geometric model capable of synthesizing natural yarn-level geometry with yarn sliding and flyaway fibers. The procedural model is driven by a specialized large language model, WeavingLLM, which is fine-tuned on an annotated dataset of formatted weaving drafts, and prompt-tuned with domain-specific fabric expertise. Through fine-tuning and prompt tuning, WeavingLLM learns to design weaving drafts and fabric parameters from textual prompts, enabling the procedural model to produce diverse weaving patterns that stick to weaving principles. The generated macro-scale texture, along with the micro-scale geometry, can be used for fabric rendering. Consequently, our framework produces materials with significantly richer detail and realism compared to prior generative models.

\end{abstract}
\section{Introduction}
\label{sec:intro}


Woven fabric materials are commonly utilized in digital human modeling, interior design, and cinematic productions. Due to their unique weaving pattern and appearance, traditional fabric material design typically involves multiple stages, including weaving pattern creation (e.g., using tools like Substance Designer), texture authoring, and shading parameter tuning. The process is time-consuming, even for experienced artists. There is a growing demand for accessible tools that allow casual users to create photorealistic fabric materials from text, without any prior knowledge of fabrics or textiles.

Diffusion models have shown impressive capabilities in generating images and even in creating materials~\cite{sartorMatFusionGenerativeDiffusion2023a, vecchioMatFuseControllableMaterial2024, vecchio2024stablematerials, vecchio2024controlmat}. Recently, it has also been introduced for text/image-guided fabric material generation~\cite{he2024dresscode, zhang2024fabricdiffusion}. While these models have significantly lowered the barrier to fabric material creation, the generated materials often produce artificial stripes, exhibit physically implausible patterns, or omit fabric microstructures, leading to unrealistic close-up renderings. The main challenges arise from the limited ability of diffusion models pretrained on natural images to generate microstructures, as well as the difficulty of applying constraints to the generation process.

In this paper, we introduce a novel framework, \emph{FabricGen}, designed to generate detailed woven fabric materials from text. This framework produces highly detailed microstructures while sticking to weaving principles, without requiring complex manual design. The core of our framework lies in the decomposition of macro-scale textures and micro-scale weaving patterns. The macro-scale texture is generated by a fine-tuned diffusion model, while the micro-scale patterns are crafted by a large language model (LLM)-driven procedural geometric model.

Since macro-scale textures need to be free of microstructures, pre-trained diffusion models alone cannot adequately produce this type of fabric. To address this, we collect a dataset of microstructure-free fabrics to fine-tune a pre-trained diffusion model, allowing us to generate a macro-scale texture/albedo map. When it comes to micro-scale weaving patterns, two key challenges arise: first, existing procedural models struggle to create natural yarn-level microstructure due to missing crucial features such as random sliding and flyaway fibers. Second, even with procedural models, experienced artists are typically required to design the weaving drafts and corresponding parameters. To tackle these challenges, we first propose an advanced procedural model for woven fabric that enhances randomness and generality. Additionally, we customize a domain-specific LLM to design weaving draft and fabric parameters. The LLM is supervised fine-tuning (SFT) on a collected weaving draft dataset and further prompt-tuned with fabric expertise to learn the weaving rule and priors of woven fabrics, enabling an end-to-end generation without any manual design or prior knowledge of woven fabrics.



The albedo generated from the fine-tuned diffusion model, along with the microstructure data (e.g., normal, tangent) produced by the procedural model, can be used for fabric rendering using a fabric shading model. Consequently, our framework is capable of creating fine-grained materials with rich microstructures, guided by only text or together with images. This demonstrates a significant improvement in quality over previous generative models that were specialized for fabric materials or general materials.

To summarize, our main contributions include:

\begin{itemize}
    \item a novel fabric generation framework--FabricGen, which decomposes macro-scale textures and micro-scale weaving patterns to enable the generation of highly detailed woven fabric materials,
    \item an LLM-driven procedural model for generating diverse micro-scale weaving patterns from text, along with multi-ply yarn geometry, yarn sliding, and flyaway fibers,
    \item a fine-tuned diffusion model for generating microstructure-free fabric textures from text/image.
\end{itemize}

\begin{figure*}[tb]
\centering
\includegraphics[width = 1.0\linewidth]{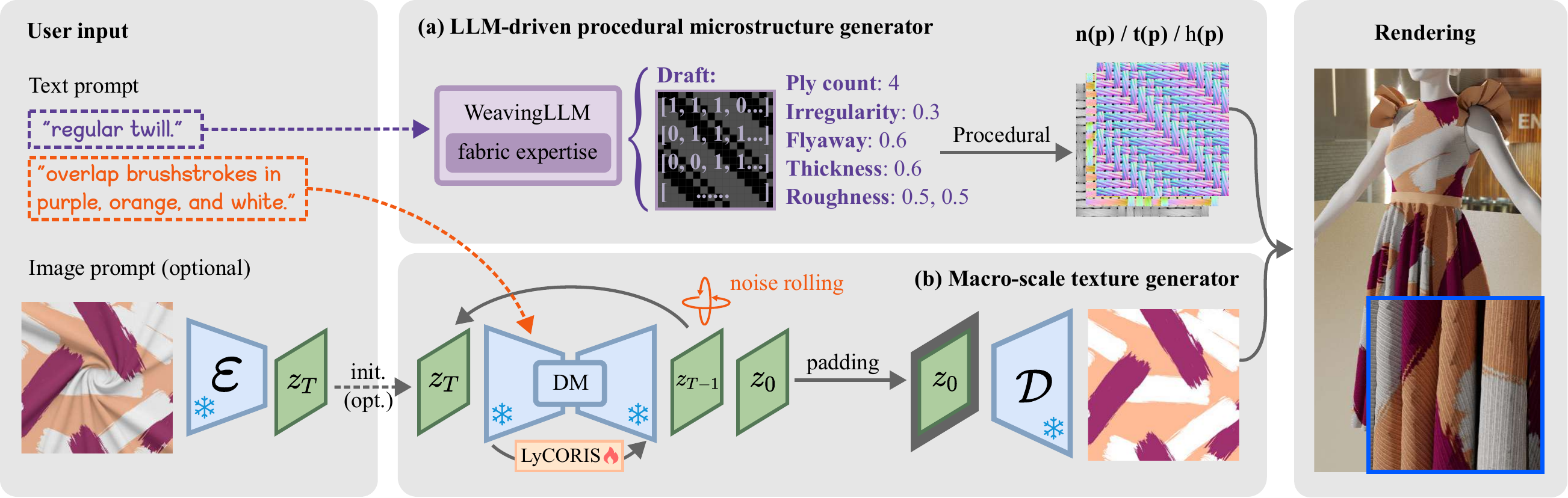}
\caption{Overview of FabricGen. Our fabric material creation framework consists of two main components: (a) an LLM-driven procedural microstructure generator that generates yarn-level microstructures along with fabric parameters, and (b) a texture generator that generates tileable, macro-scale textures from a text prompt (and optional image). \revise{Note that the two components require distinct prompts.}}
\label{fig:pipeline}
\end{figure*}

\section{Related Work}
\label{sec:related}

\paragraph{Material generation}
Material generation has emerged as an active research area, aiming to synthesize plausible textures from images or text prompts~\cite{bergmannLearningTextureManifolds2017,zhouTileGenTileableControllable2022a,rodriguez-pardoSeamlessGANSelfSupervisedSynthesis2023b,vecchio2024controlmat,vecchioMatFuseControllableMaterial2024,vecchio2024stablematerials,zhang2024fabricdiffusion}.  
Early works primarily adopted generative adversarial networks (GANs) to model high-frequency details~\cite{bergmannLearningTextureManifolds2017,zhouTileGenTileableControllable2022a,rodriguez-pardoSeamlessGANSelfSupervisedSynthesis2023b}, e.g., TileGen~\cite{zhouTileGenTileableControllable2022a} combined tileable StyleGAN2 variants~\cite{karrasAnalyzingImprovingImage2020} with conditional patterns to reconstruct artifact-free SVBRDF from single images under material category constraints.  
Recent approaches increasingly leverage diffusion models~\cite{ying2025chord,xue2024reflectancefusion,ma2025materialpicker,vecchio2024controlmat,vecchioMatFuseControllableMaterial2024,vecchio2024stablematerials}.  
ControlMat~\cite{vecchio2024controlmat} formulates SVBRDF reconstruction as conditional synthesis, generating tileable high-resolution materials from uncontrolled photos.  
MatFuse~\cite{vecchioMatFuseControllableMaterial2024} enables controllable generation and editing of 3D materials via multi-modal conditioning and latent manipulation.  
MaterialPicker~\cite{ma2025materialpicker} employs a Diffusion Transformer for text or photo-guided generation with distortion correction.  
ReflectanceFusion~\cite{xue2024reflectancefusion} adopts a tandem diffusion framework for text-driven editable SVBRDF.  
Generative neural materials~\cite{GNNRaghavan} produce BTFs from text or images via a universal NeuMIP basis.  
Chord~\cite{ying2025chord} utilizes two-stage diffusion and decomposition for high-quality SVBRDF generation.  
While these diffusion-based methods have significantly advanced general material generation, they still face common limitations: the inability to enforce domain-specific structural rules (e.g., weaving principles in fabrics), difficulty in producing realistic yarn-level microstructures, and resolution constraints that hinder fine-grained close-up rendering.

\paragraph{Fabric material generation}
Recently, there are some methods that focus on fabric generation~\cite{zhang2024fabricdiffusion,he2024dresscode}. FabricDiffusion~\cite{zhang2024fabricdiffusion} formulates texture transfer as tileable material generation via diffusion, enabling distortion-free PBR textures to be generated from a single clothing image. DressCode~\cite{he2024dresscode} employs a fine-tuned Stable Diffusion model with three distinct decoders to synthesize tile-based PBR textures (diffuse, normal, and roughness maps) from text prompts.

While these methods are robust to surface variations and enable diverse outputs, their resolution limitations often lead to blurred details and structural artifacts. Different from the above works, we propose a new approach: the weaving pattern is generated separately from the color pattern by an LLM-driven procedural model, which ensures both pattern diversity and fine fabric details. \revise{To the best of our knowledge, no existing work built for text-to-procedural microstructure generation.}

\paragraph{Procedural fabric model}

Prior works have proposed various procedural geometric models for woven fabrics. One class of methods models fabrics at the yarn, ply, or even fiber level using explicit curves~\cite{Zhao2016fittingyarn, Schröder2015Reverse, Montazeri2020ply, zhu2023yarn, Li2024Fiber, Wu2017realtime}. These approaches achieve photorealistic results but rely on explicit geometric representations at the ply or even fiber level, incur computational and storage costs during both data generation and rendering.
In contrast, another class of surface-based methods~\cite{IrawanAndMarschner2012, Jin:2022:inverse} treats the fabric as a 2D sheet and procedurally generates spatially-varying normals, tangents, and heights for rendering. By avoiding explicit geometry construction and intersection tests, surface-based models are highly efficient and integrate seamlessly with traditional rendering pipelines.

However, existing surface-based models have notable limitations: they typically assume single-ply yarns and neglect global irregularities, such as yarn sliding and flyaway fibers that contribute to natural fabric irregularity. In this paper, we propose an enhanced procedural geometric model to handle these issues.




\begin{figure}[tb]
\centering
\includegraphics[width = \linewidth]{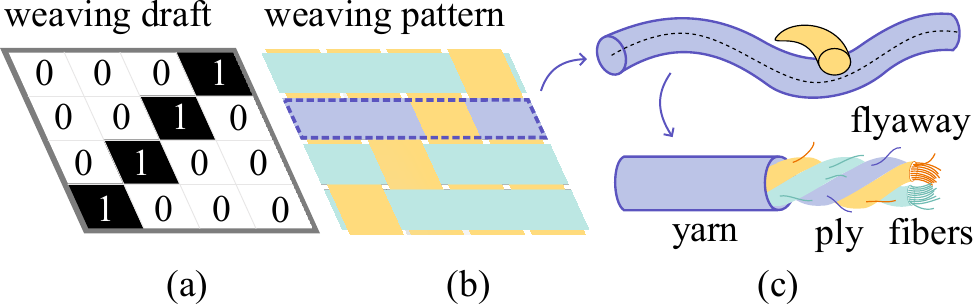}
\caption{
Woven fabrics are typically constructed by interlacing weft and warp yarns following specific patterns (b), which can be represented via a binary matrix, named weaving draft (a). Each yarn may consist of multiple plies, and each ply can be further divided into a number of fibers. Some fibers escape from the surface, resulting in irregular flyaway fibers (c).}
\vspace{-15pt}
\label{fig:composition}
\end{figure}

\section{Background}
\label{sec:background}


\paragraph{Microstructure of woven fabric}
Woven fabrics are typically composed of horizontal and vertical yarns (namely, wefts and warps) interlaced in specific weaving patterns. The weaving pattern within a repeat unit can be represented as a binary matrix, referred to as a weaving draft. The diverse configurations of warp and weft yarns allow for the creation of a wide variety of appearances~\shortcite{HandWeaving.net}, which are typically designed by skilled artists. Previous works~\cite{Jin:2022:inverse,Tang2024Woven} typically focus on several classical patterns (twill, satin, and plain), ignoring the diversity implied in weaving patterns. In this paper, we first propose to generate weaving patterns directly from natural language descriptions.

As for yarn, each yarn may consist of a single strand or multiple strands (namely, plies), and each ply is composed of numerous fine fibers that are twisted together. Beyond these regular patterns, there are also some irregular features in the fabric, such as flyaway fibers escaping from the surface and yarns occasionally slipping out of place. We illustrate the basic composition of woven fabric in Fig.~\ref{fig:composition}.


\paragraph{Procedural yarn representation}
Existing methods~\cite{IrawanAndMarschner2012, Jin:2022:inverse} model yarns as curved cylinders, where the yarn centerline is represented as a circular arc parameterized by the maximum inclination angle $u_{\mathrm{max}}$, and the yarn cross-section is assumed to be circular, with individual fibers twisted around the yarn surface at a specified twist angle $\psi$. 

The model analytically projects the 3D yarn geometry on the 2D UV domain. Formally, the procedural model $\mathcal{F}$ can be defined as a set of geometric functions over UV space:
\begin{equation}
    \mathcal{F} = \left\{ \mathbf{n}(\mathbf{p}),\ \mathbf{t}(\mathbf{p}),\ h(\mathbf{p}) \right\}, \quad \mathbf{p} = (x,y) \in [0,1]^2,
\end{equation}
where $\mathbf{p}$ denotes the UV coordinate, $\mathbf{n}$  the surface normal, $\mathbf{t}$ the tangent direction (or namely, orientation), and $h$ the relative height field. Together, these functions define a parameterization of the yarn-level microstructure over the UV domain, enabling seamless integration with traditional rendering pipelines and allowing direct export of derived maps (e.g., normal, height, or tangent maps).

\section{FabricGen: a two-scale fabric material generation framework}
\label{sec:overview}

In this paper, we propose FabricGen, a novel fabric material generation framework, which generates high-quality fabric materials in an end-to-end manner given natural language prompts. At the core of the framework is the decoupling of the macro-scale fabric texture/albedo generation (Sec.~\ref{sec:macro}) and the micro-scale fabric microstructure generation (Sec.~\ref{sec:micro}), where the former is generated by a fine-tuned diffusion model and the latter is generated via an LLM-driven procedural geometric model. The diffusion model allows the diverse creation of macro-scale fabric textures, and the LLM-driven procedural model creates diverse and detailed microstructures.
Finally, we employ fusion rendering using SpongeCake~\cite{Wang:2023:SpongeCake}, \revise{a layered shading model. Rendering details are provided in the supplementary material.} As a result, our framework produces materials that exhibit both visual diversity and fine-grained structural detail. An overview of our framework is shown in Fig.~\ref{fig:pipeline}.




\subsection{Macro-scale fabric texture generation}
\label{sec:macro}


Previous works~\cite{he2024dresscode,zhang2024fabricdiffusion} leverage diffusion models to generate fabric materials, by fine-tuning the pre-trained diffusion model with a collection of fabric materials. Their generated materials are supposed to include both the texture and the fabric microstructure together. Unfortunately, due to resolution limitations and the lack of structural constraints, the resulting materials often violate weaving rules or entirely omit microstructural details. To address these issues, we constrain the generative model to produce pure albedo maps that exclude microstructures. To this end, we collect a dataset of microstructure-free fabric textures and then fine-tune the pre-trained diffusion model with this dataset to generate albedo maps. In addition to text prompts, the generative model also supports optional fabric image conditions. Additionally, the generated materials should be seamless.


\paragraph{Diffusion model fine-tuning} 
We collect a dataset of 600 microstructure-free fabric textures along with the corresponding description. Using the collected dataset, we fine-tune the FLUX.1-dev~\cite{flux2024} model with LyCORIS~\cite{yeh2023navigating}, adapting its output to directly produce microstructure-free albedo maps and converting a general-purpose image generator into a fabric-specific albedo generator. We incorporate noise rolling mechanism~\cite{vecchio2024controlmat} to ensure tileability. Training details are shown in the supplementary.

\paragraph{Multimodal conditioning} Our model also supports multimodal conditioning by inputting fabric images, either flat or with wrinkles. The input image is encoded into the latent space via a VAE decoder, and the resulting latent vector is used as an initialization condition for the diffusion model. This guides the generation process to suppress geometric wrinkles while preserving the original pattern style. 

\subsection{LLM-driven procedural microstructure generation}
\label{sec:micro}

Besides the macro-scale texture, the key to fabrics is the microstructure, consisting of several levels (weaving pattern, yarn, ply, and fiber), as described in Sec.~\ref{sec:background}. Generating these hierarchical structures is challenging for diffusion models~\cite{he2024dresscode, zhang2024fabricdiffusion} due to their limited output resolution. To this end, we resort to the procedural model for microstructure generation and leverage a LLM for weaving draft creation. Unfortunately, existing surface-based procedural approaches~\cite{Jin:2022:inverse, IrawanAndMarschner2012} generally neglect global irregular features, and their single cylinder model does not support multi-ply configuration. To handle these issues, we propose a novel procedural geometric model with enhanced expressiveness and natural irregularity. 


\paragraph{Procedural geometric model}
Our procedural model defines the yarn-level microstructures given a weaving pattern draft and a set of yarn parameters.
We model plies in a yarn as curved helices, enabling multi-ply configuration, as shown in Fig.\ref{fig:definition_yarn}. We also introduce some global irregular features, including yarn sliding and flyaway fibers. All these components constitute our final procedural geometric model, which produces spatially-varying geometric data, such as normal, orientation, and height (as visualized in Fig.~\ref{fig:visual_multiply}). Notably, our model does not require explicit generation of these maps. Instead, it supports on-demand queries, enabling ultra-fine details without the need for precomputing or storing high-resolution textures. Further details of our procedural model are provided in Sec.~\ref{sec:procedural}.

\paragraph{Weaving draft generation} With the proposed fabric procedural model, it's still not clear how to design the weaving pattern and geometric parameters automatically. Therefore, we propose WeavingLLM, a fine-tuned LLM that learns to generate formatted weaving drafts and corresponding parameters (e.g., roughness and ply counts) from natural language prompts, just like an experienced designer. The model is fine-tuned from the pre-trained Qwen2.5-14B-it~\cite{qwen2.5} model using QLoRA~\cite{dettmers2023qlora} on a collected dataset of 1,142 annotated weaving drafts sourced from Handweaving.net~\cite{HandWeaving.net}. All drafts are constrained to a maximum size of 16×16. In addition to fine-tuning, we apply prompt tuning with domain-specific fabric expertise, enabling LLM to learn the prior knowledge of various types of fabrics.

Given a text prompt, WeavingLLM first generates a binary matrix as the weaving draft. The LoRA adapter is then disabled, and the base LLM predicts the fabric parameters following the fabric expertise. The weaving draft and parameters are subsequently used for procedural microstructure synthesizing and rendering. The LLM instructions, parameters list, and fine-tuning details are provided in the supplementary material.

Since we only generate the weaving pattern within a repeat unit, our model inherently adheres to fundamental weaving principles and produces rich details. The generated weaving drafts are constrained to a maximum size of 16×16; therefore, in theory, a loom with 16 shafts and treadles can physically realize any output of our model.

\section{Procedural geometric model}
\label{sec:procedural}

In this section, we introduce our procedural geometric model with several novel components. We first introduce the curved helix model for multi-ply yarns. Then, we discuss how to model natural irregularities. 

\subsection{Curved helix model}
\label{sec:procedural:definition}


\begin{figure}[tb]
\centering
\includegraphics[width = 1.0\linewidth]{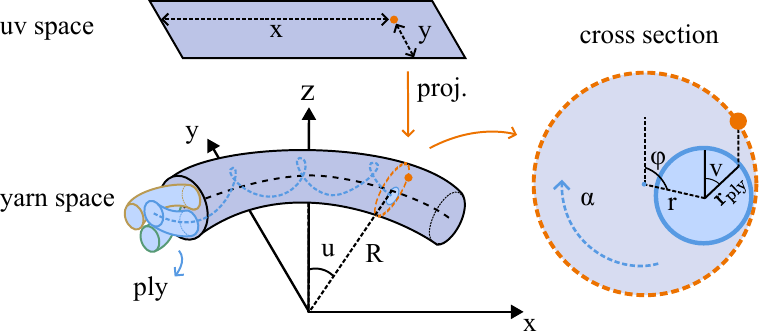}
\caption{ Curved helix model for multi-ply yarns. The model achieves analytical normal/orientation formulation, where the yarn space arc parameters $u, v$ are linear mapped from UV space. }
\vspace{-10pt}
\label{fig:definition_yarn}
\end{figure}

Previous works~\cite{Jin:2022:inverse, IrawanAndMarschner2012} model yarns as curved cylinders, which only support single-strand yarn configuration. Wu et al.~\cite{Wu2017realtime} model plies as helix curves. However, it requires explicit storage and rendering of each curve. Instead, we propose a procedural curved helix yarn model to map ply-level geometry directly from UV coordinates, depicted in Fig.~\ref{fig:definition_yarn}.
We model yarn centerline as a circular arc, while individual plies form helices around the yarn centerline with rotation speed $\alpha$. We parameterize the plies through yarn space arc coordinates $u$ and $v$, which are linearly projected from surface UV coordinates $x, y$. Therefore, we drive the analytical normal $\mathbf{n}$, orientation $\mathbf{t}$ and height $h$:

\begin{eqnarray}
    \mathbf{n}(u,v) &=& 
    \begin{bmatrix}
    \sin(u)\cos(v) \\
    \sin(v) \\
    \cos(u) \cos(v)
    \end{bmatrix}, \\
    \mathbf{o_\mathrm{ply}}(u) &=& 
    \begin{bmatrix}
    \cos(u) \\
    -r \alpha \cos(\varphi(u)) \\
    -\sin(u) - r \alpha \sin(\varphi(u))
    \end{bmatrix}, \\
    \mathbf{t}(u, v) &=& \mathrm{rotate}(\mathbf{o_{\mathrm{ply}}}, \mathbf{n}, \psi) \\
    h(u,v) &=& r\cos(\varphi(u)) + \cos(u)(R + r_{\mathrm{ply}}\cos(v)), \quad \\
    \varphi(u) &=& \varphi(0) + uR\alpha.
\end{eqnarray}
where $R$ is the radius of the yarn arc, $r_\mathrm{ply}$ is the ply radius, $r$ is the distance from the ply center to the yarn center, $\varphi (u)$ indicates rotational phase of ply along the axis, $\varphi(0)$ is the initial phase of the ply. The fiber orientation $\mathbf{t}$ comes from rotating the ply orientation $\mathbf{o_{\mathrm{ply}}}$ around the normal at a certain twist angle $\psi$. An visualization of these geometric functions is shown in Fig.~\ref{fig:visual_multiply}. More derivation and configuration details are provided in the supplementary material.


\begin{figure}[tb]
\centering
\includegraphics[width = 0.9\linewidth]{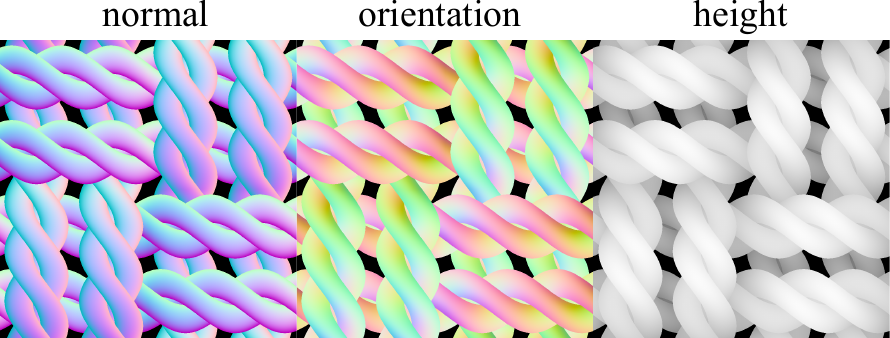}
\caption{ Visualization of the normal, orientation and height maps of a 3-ply basket weave. \revise{The parameters of this case are provided in the supplementary material.} }
\label{fig:visual_multiply}
\vspace{-25pt}
\end{figure}

\begin{figure*}[tb]
\centering
\includegraphics[width = \linewidth]{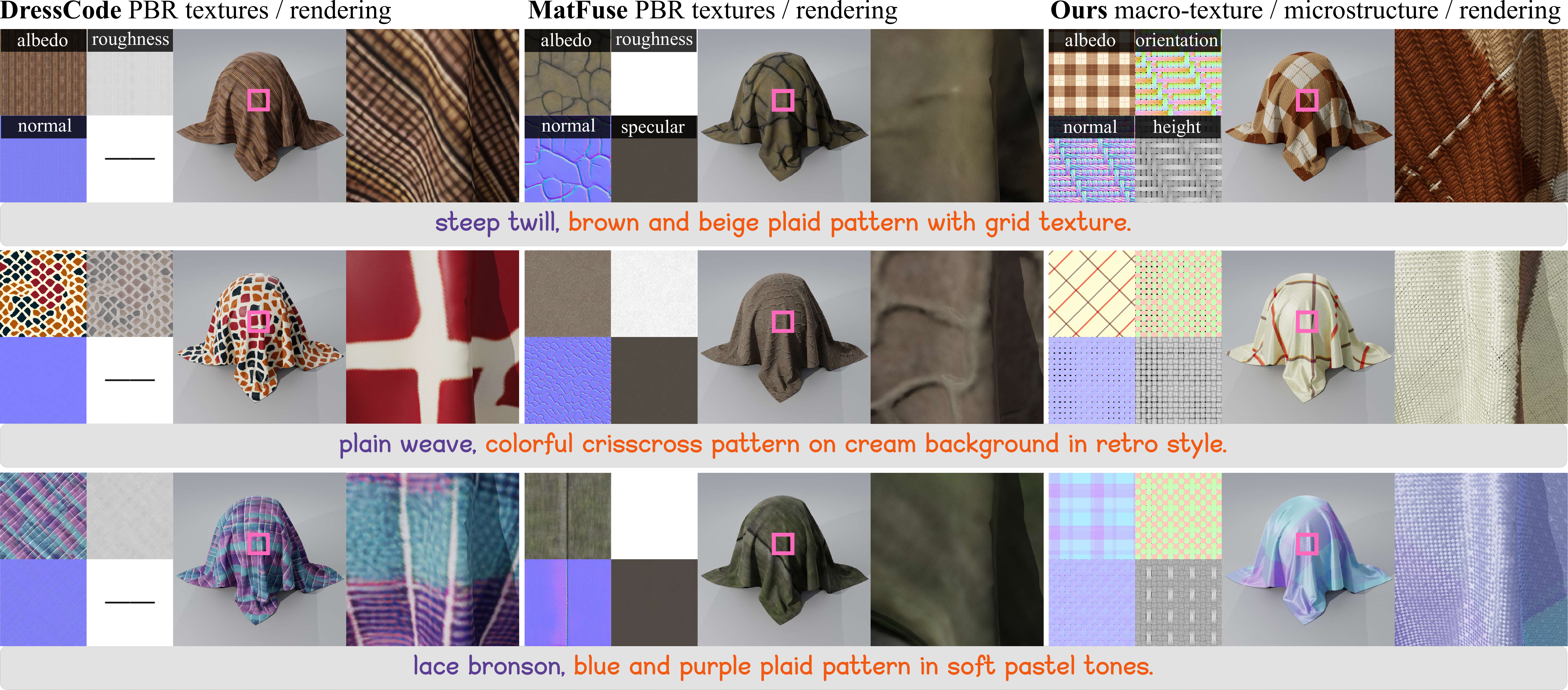}
\caption{\textbf{Comparison on text conditioning.} Given text prompt, DressCode~\shortcite{he2024dresscode} and MatFuse~\shortcite{vecchioMatFuseControllableMaterial2024} generate PBR textures, which struggle to capture
fine fabric details, resulting in blurring and distortion in close-up views. In contrast, our method generates procedural microstructure together with macroscale texture, ensuring a realistic appearance in both macro and close-up views.}
\label{fig:result_compare}
\end{figure*}

\subsection{Global irregular effects}
\label{sec:procedural:irregular}

To model natural irregularities of woven fabric, existing methods~\cite{IrawanAndMarschner2012, Jin:2022:inverse} apply perturbations to the specular reflectance or the height field, thereby achieving a macroscopic irregular appearance. However, they neglect essential global irregular effects, such as yarns sliding out of place and fibers escaping from the surface. In this section, we discuss how to simulate these phenomena.


\paragraph{Yarn sliding} The yarn arrangement in real fabrics is not always regular but exhibits inherent positional randomness, manifesting as yarn sliding. Thus, we model this effect by perturbing yarn positions using continuous procedural noise. Given a UV point $\mathbf{p} = (x,y)$ following the coordinate system in Fig.~\ref{fig:definition_yarn}, we introduce axial ($x$-direction) 1D Perlin noise $P(x)$ on the yarn, then apply radial ($y$-direction, $y\in(0,1)$) perturbations through:
\begin{eqnarray}
    y_s &=& 0.5 + (y-0.5) (1 - k_{\mathrm{sliding}}|P(x)|), \\
    y_r &=& y_s^{e^{k_{\mathrm{sliding}}P(x)}},
\end{eqnarray}
where $y_s$ denotes the scaled position, $y_r$ is the final perturbed coordinate, and $k_{\mathrm{sliding}}$ is a user-defined parameter that controls the sliding strength. The bijective mapping $f:y\rightarrow y_r$ enables practical implementation through inverse queries to retrieve regular coordinates from irregular space. This sliding mechanism reveals otherwise occluded lower-layer yarns, with visual plausibility guaranteed by our procedural model.

\paragraph{Flyaway fibers} 
Flyaway refers to fibers escaping from the fabric surface. To model this, we propose to introduce an additional fiber layer with stochastic fiber orientations in the SpongeCake model, where the 3D fiber orientation field $\mathbf{o_{flyaway}}(\mathbf{p})$ is constructed from two 2D Perlin noise $N_1, N_2$, where $N_1$ controls the fiber position and horizontal orientation, and $N_2$ controls the vertical orientation. Though carefully design, our stochastic orientation field exhibits a continuous fiber-like distribution. Detailed configuration is provided in the supplementary material.

    




    
    

\begin{table}[tb]
    \caption{\label{tab:cmp_text} Quantitative comparisons between our method, MatFuse~\shortcite{vecchioMatFuseControllableMaterial2024}, and DressCode~\shortcite{he2024dresscode}.}
    \centering
    \begin{tabular}{c c c c}
    \toprule
    \quad & MatFuse & DressCode  & Ours \\

    \midrule
    & \multicolumn{3}{c}{\textbf{Text conditioning}} \\
    \midrule
    CLIP Score $\uparrow$  & 0.240 &   0.307 &  \textbf{0.317} \\ 
    User study $\uparrow$  & 1.43\% &   16.02\% &  \textbf{82.55\%} \\ 

    \midrule
    & \multicolumn{3}{c}{\textbf{Text and image conditioning}} \\
    \midrule

    CLIP-I Score $\uparrow$  &  0.722 & N/A & \textbf{ 0.827} \\ 
    
    \bottomrule
    \end{tabular}
\end{table}


\section{Results}
\label{sec:results}

In this section, we first present our fabric material generation results and compare them with two representative text-to-material methods: DressCode~\cite{he2024dresscode}, which targets fabric material generation, and MatFuse~\cite{vecchioMatFuseControllableMaterial2024}, which focuses on general materials. Then we present our results with image conditioning, and compare them with MatFuse. Finally, we perform ablation studies on several key components of our pipeline. All training, inference, and rendering are executed and evaluated on an NVIDIA RTX 4090 GPU.

\begin{figure}[tb]
\centering
\includegraphics[width = \linewidth]{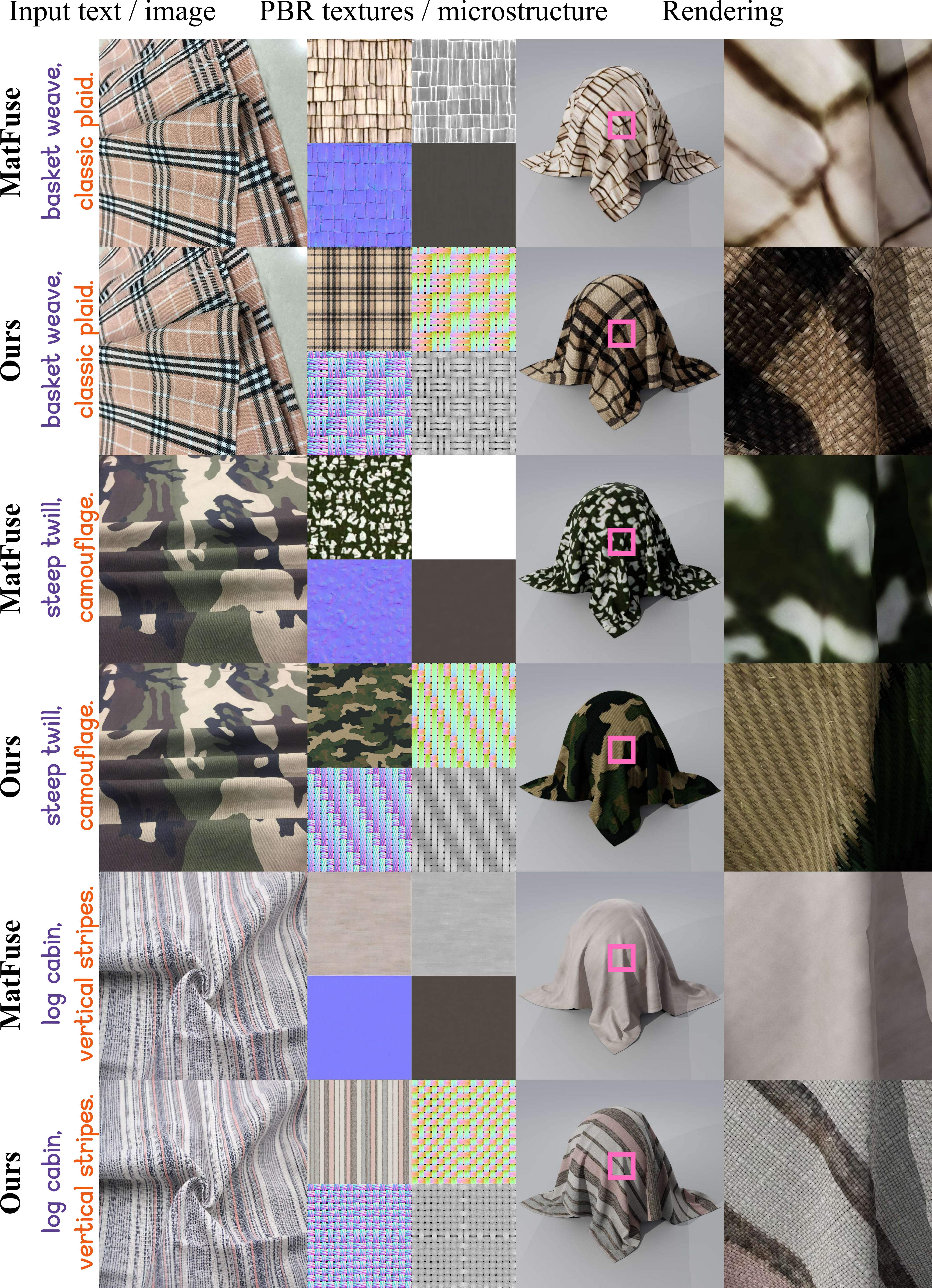}
\caption{ \textbf{Comparison on text and image conditioning}. Given both text and image prompts, our generated materials achieving better consistency and richer detail than MatFuse~\cite{sartorMatFusionGenerativeDiffusion2023a}. }
\label{fig:result_img2img_0}
\vspace{-15px}
\end{figure}

\begin{figure}[tb]
\centering
\includegraphics[width = \linewidth]{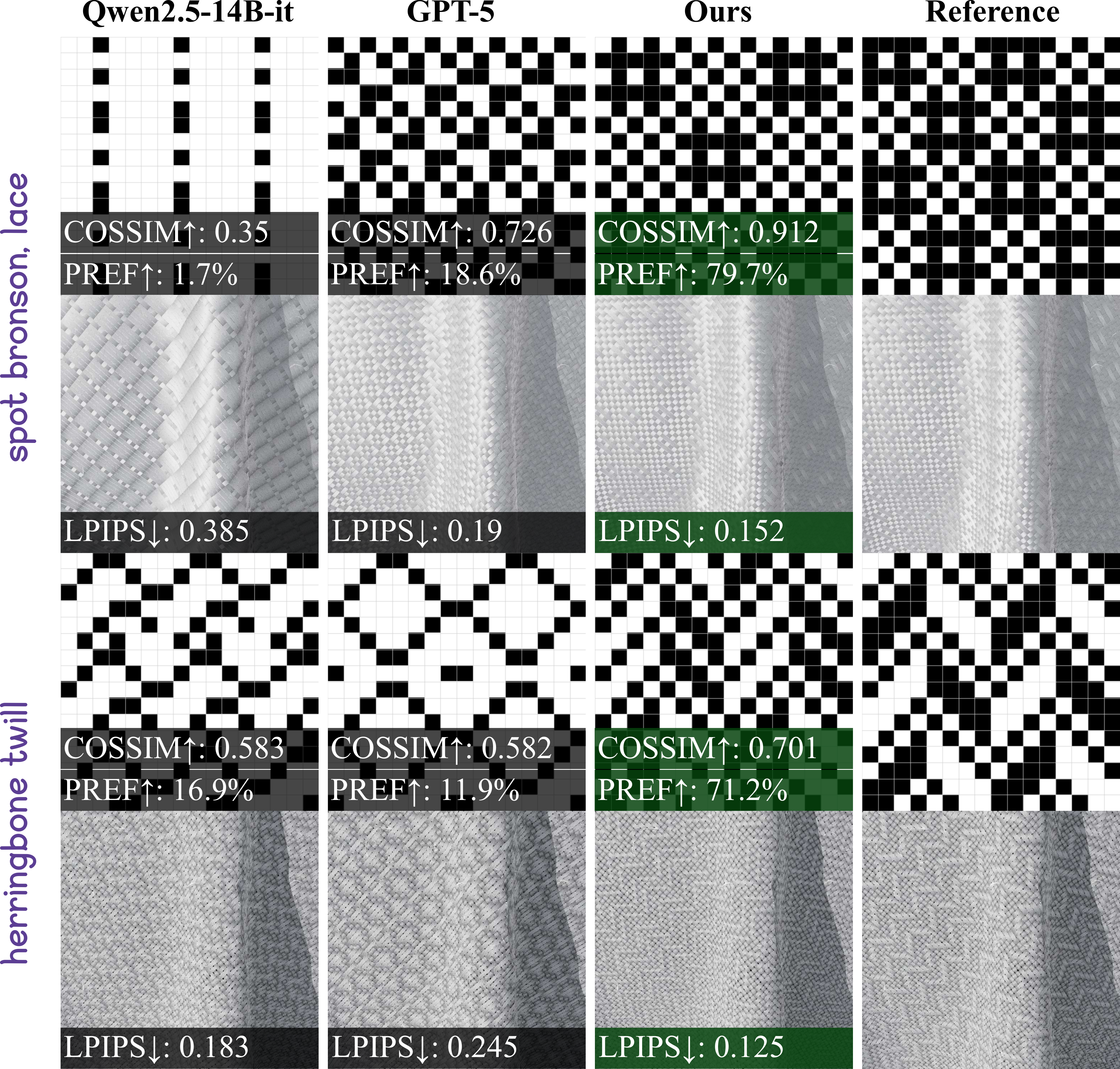}
\caption{ \textbf{Comparison among WeavingLLM, Qwen2.5~\cite{qwen2.5}, and GPT-5~\cite{openai_gpt5_2025}}. COSSIM indicates the cosine similarity of Fourier spectra, and PERF indicates the user preferences ratio. All renderings use a pure gray albedo map. More reference drafts are provided in the supplementary material.}
\label{fig:ablation_pattern_0}
\vspace{-5px}
\end{figure}

\begin{figure}
\centering
\includegraphics[width = 1.0\linewidth]{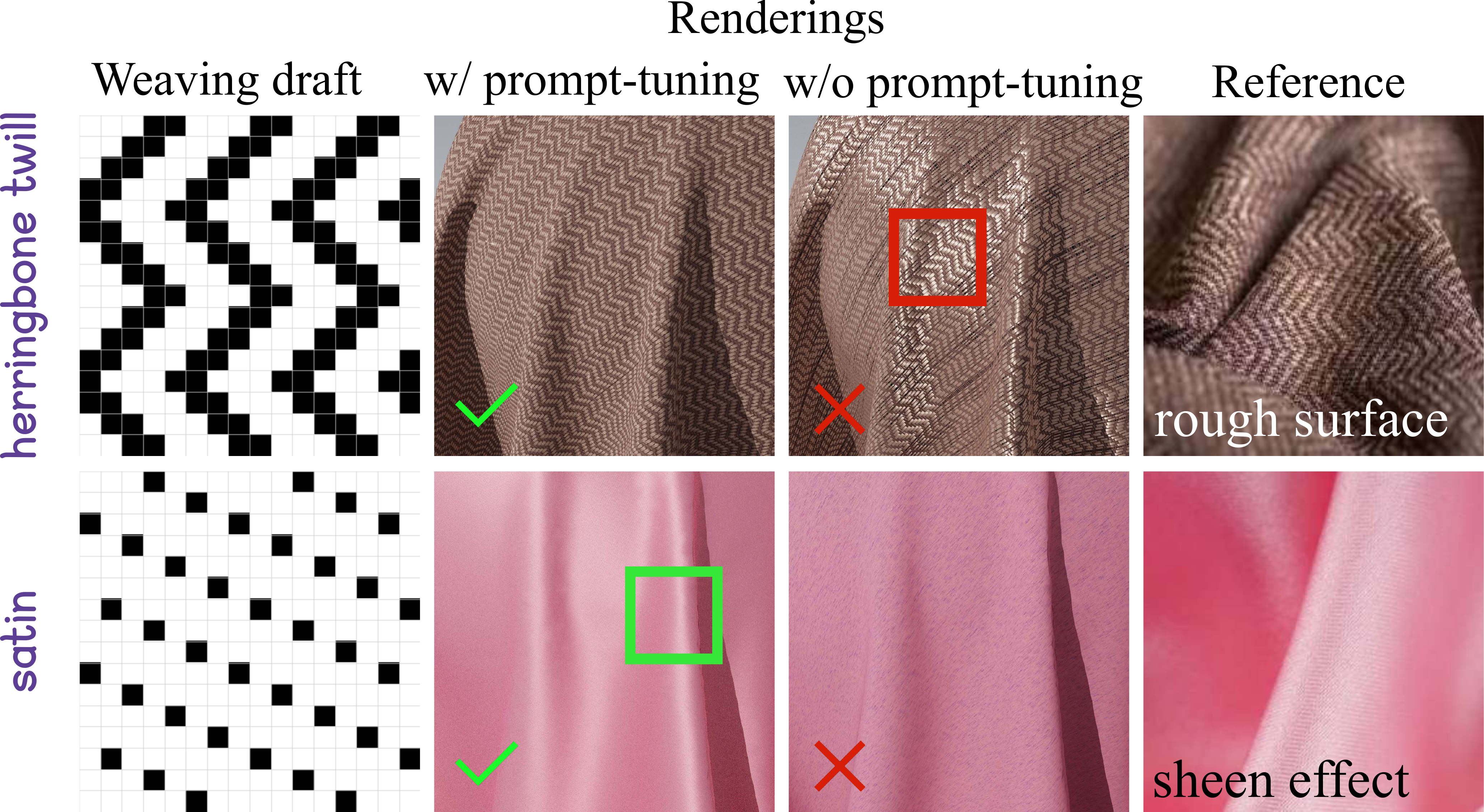}
\caption{ \textbf{Impact of prompt-tuning}. The parameters predicted by WeavingLLM successfully reproduce the rough appearance of herringbone twill and the anisotropic sheen of satin, whereas parameters predicted by the base LLM fail to reproduce this effect.}
\label{fig:ablation_params}
\vspace{-10pt}
\end{figure}

\begin{figure}[tb]
\centering
\includegraphics[width = 1.0\linewidth]{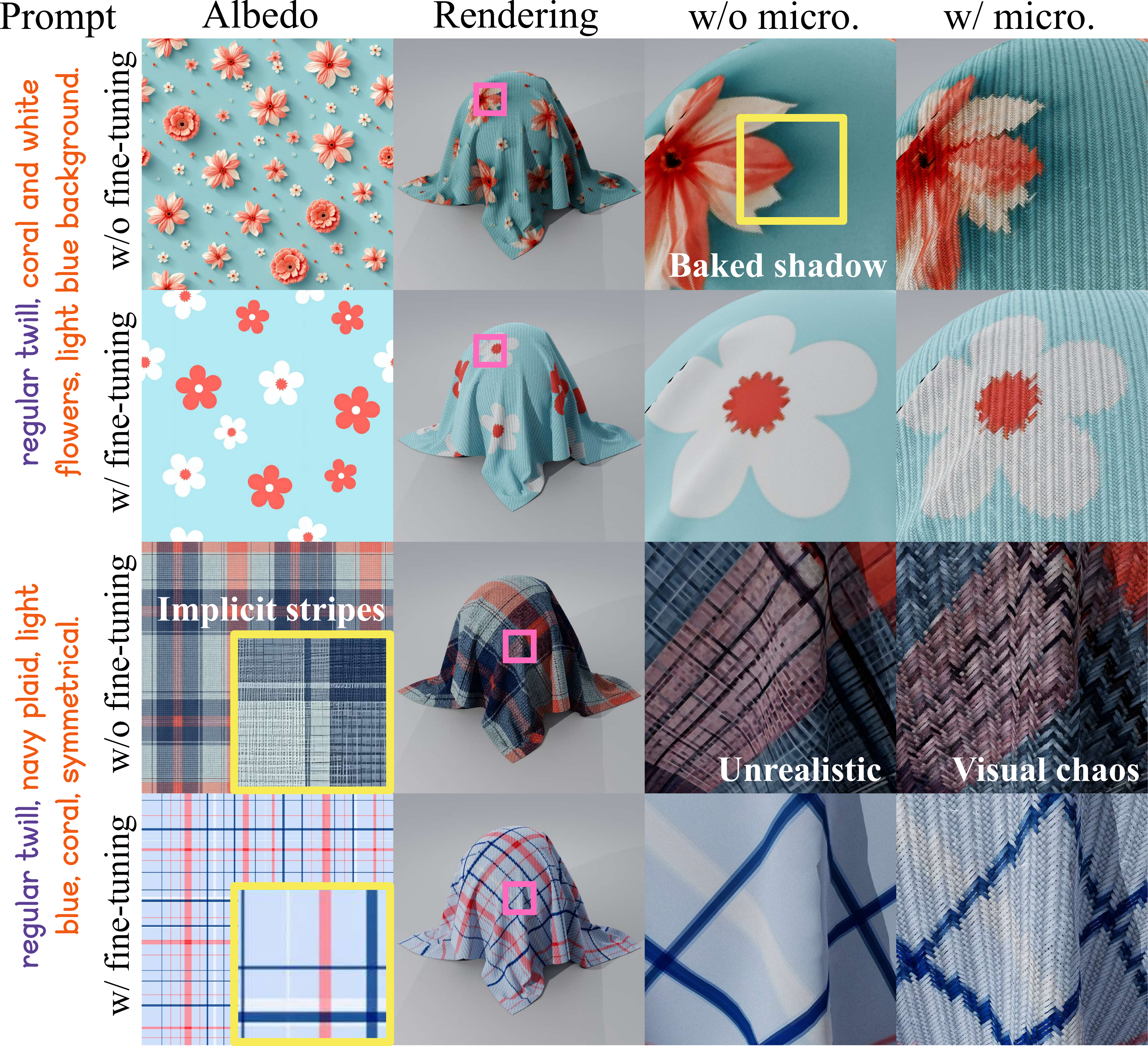}
\caption{ \textbf{Impact of macro-scale texture generator}. The diffusion model pre-trained on natural images occasionally produces unintended 3D artifacts (first row) or implicit stripes (third row), whereas our fine-tuned model is capable of generating microstructure-free albedo maps. The results in the fourth column use the regular twill pattern in Fig~\ref{fig:pipeline}.}
\label{fig:ablation_flux}
\end{figure}

\begin{figure}[tb]
\centering
\includegraphics[width = 1.0\linewidth]{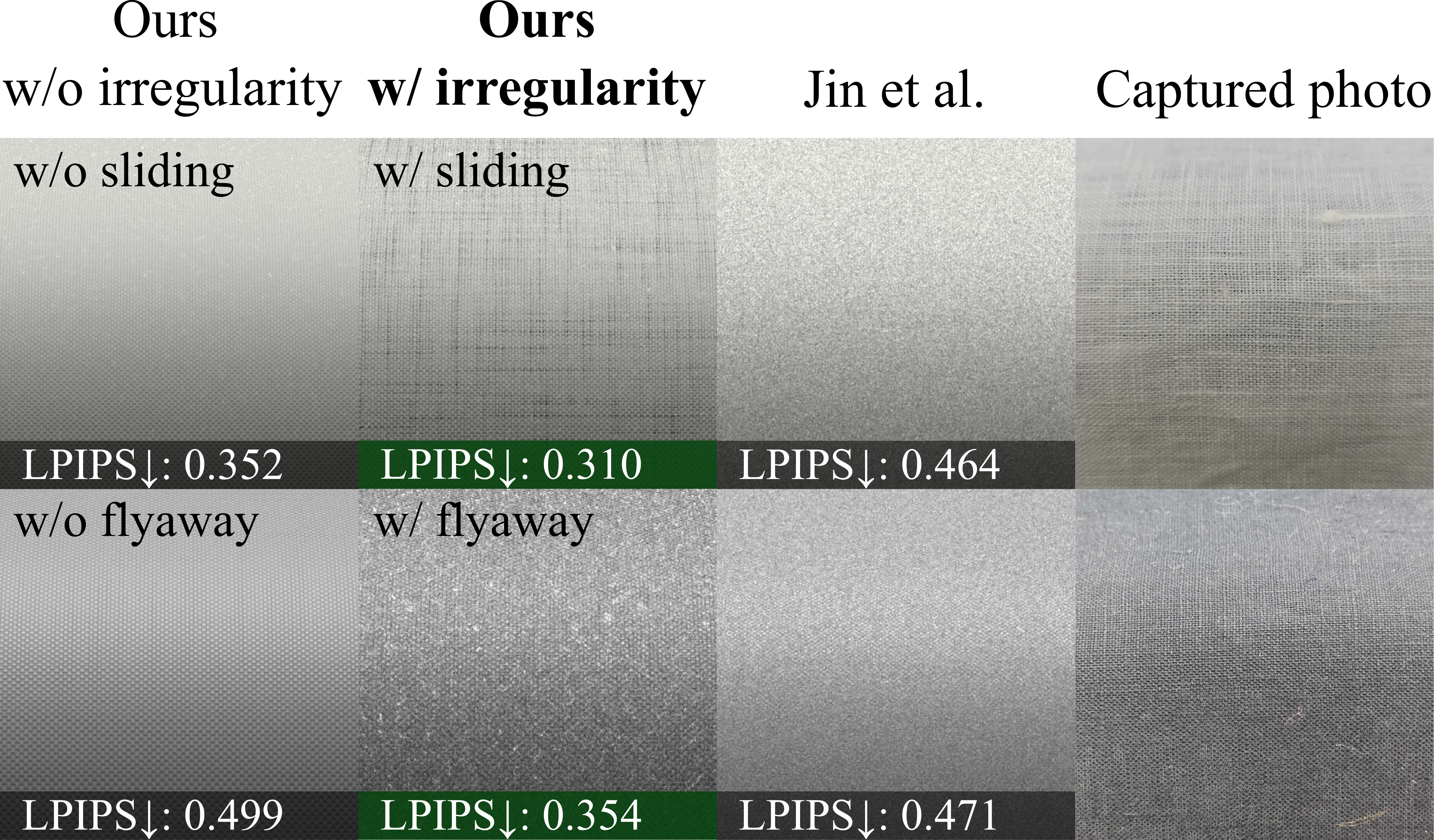}
\caption{ \textbf{Impact of irregular effects}. Irregular yarn displacements in real fabrics often result in strip-like gaps between yarns, and fibers escaping from the surface cause irregular highlights.}
\label{fig:ablation_irregular}
\vspace{-10px}
\end{figure}

\subsection{Fabric material generation}
\label{sec:generated}

\paragraph{Text conditioning} We compare our method with DressCode and MatFuse in Fig.~\ref{fig:result_compare}. Given text prompts, both prior methods generate PBR texture maps, while ours produces macro-scale textures and micro-scale weaving patterns (visualized as
normal, orientation, and height map). By rendering the generated materials on a cloth mesh, existing methods yield plausible results at a distance but suffer from blurring and distortion under close-up inspection. In contrast, our method faithfully preserves fine-scale fabric details, delivering high visual fidelity at both the macro and micro levels. We quantitatively evaluate our results using the CLIP Score~\cite{hesselCLIPScoreReferencefreeEvaluation2022} and a user study. The CLIP Score is computed as the average over 100 generated materials from diverse text prompts, measured between each prompt and plane rendering. The user study includes 12 cases, in which 64 participants are asked to choose the result that best matches the prompt and appears most realistic. As shown in Tab~\ref{tab:cmp_text}, our method achieves higher CLIP Scores and receives better user preference, demonstrating better semantic alignment, perceptual quality, and visual realism. More results covering a wide range of patterns and examples of our user study are provided in the supplementary material.

\vspace{-10pt}




\paragraph{Image conditioning} In addition to text-based generation, our method also supports additional image conditioning, providing enhanced controllability over the generated materials. We compare our method with MatFuse~\cite{vecchioMatFuseControllableMaterial2024} on the same text and image conditions, as shown in Fig.~\ref{fig:result_img2img_0}. To evaluate the results, we compute the average CLIP-I Score~\cite{Radford2021LearningTV} between the rendering result and input image, as shown in Tab.~\ref{tab:cmp_text}. The results demonstrate that our method matches better with the input image. \revise{Although FabricDiffusion~\cite{zhang2024fabricdiffusion} is not included since it is not fully open-sourced, it shares a similar scheme (Diffusion-based generation) and limitation (lack of yarn-level details) with MatFuse.} More results are provided in the supplementary material.



\subsection{Ablation studies}
\label{sec:ablation}

\paragraph{Impact of WeavingLLM} We first validate the necessity of fine-tuning a LLM in Fig.~\ref{fig:ablation_pattern_0}, by comparing WeavingLLM and our base model, Qwen2.5~\cite{qwen2.5}, and GPT-5~\cite{openai_gpt5_2025}. To quantitatively evaluate the quality of the generated weaving pattern, we find several reference drafts in the dataset that match the description, and then compute the average cosine similarity of the Fourier spectra between the generated draft and all reference drafts. In addition, we conduct a user study, inviting 60 participants to select the best-matching pattern based on the given prompt and reference examples. Moreover, we compute the learned perceptual image patch similarity (LPIPS)~\cite{zhang2018lpips} between the rendering of generated draft and reference draft. The results demonstrate that the drafts generated by WeavingLLM match better with the references, both in terms of quantitative analysis and user perception. More results cover a wide range of patterns are provided in the supplementary material.


Furthermore, we illustrate the necessity of prompt-tuning the LLM in Fig.~\ref{fig:ablation_params}. Since each type of pattern has a specific characteristic, parameters generated by an LLM without domain-specific priors may lead to an unrealistic appearance that contradicts daily experience. In contrast, with the given fabric expertise and priors, WeavingLLM is able to predict reasonable parameters according to the pattern characteristics.

\paragraph{Impact of macro-scale texture generator} We illustrate the necessity of fine-tuning the diffusion model in Fig.~\ref{fig:ablation_flux}. The original FLUX model occasionally generates 3D objects or implicit stripes, even if we have already instructed the model to output ``no microstructure, no folds, no wrinkles'' textures. These implicit details result in baked shadows and visually chaotic appearances. In contrast, our fine-tuned model produces microstructure-free fabric textures, along with the procedural microstructure, yielding cleaner and more stable rendering results.

\paragraph{Impact of our irregular effects} We validate the influence of yarn sliding and flyaway in Fig.~\ref{fig:ablation_irregular}. By introducing yarn sliding effect, our rendering matches the strip-like gaps of real fabric. By introducing flyaway fibers, we reproduce the stochastic specular highlights captured in real photo. In contrast, the previous method~\cite{Jin:2022:inverse} only applies simple perturbations on the color and height field, which are insufficient to capture these complex effects. To quantitatively validate our method, we compute the LPIPS between the renderings and captured photos, demonstrating that our method matches better with the irregularity of real fabrics.

\subsection{Discussion and limitations}
\label{sec:discussion}

Our approach currently focuses on printed woven fabrics. However, there exist other types of fabrics, such as knitted fabrics, plastisol print, \revise{jacquard}, and embroidery. Extending support to more diverse types is left as future work. 



\section{Conclusion}
\label{sec:conclusion}

We introduce FabricGen, an end-to-end framework for generating high-fidelity woven fabric materials from text/image prompts. By decoupling macro-scale texture synthesis from micro-scale geometry modeling, our method produces photorealistic and semantically accurate fabric materials. At the macro level, we fine-tune a diffusion model to produce seamless, microstructure-free textures. At the micro level, a procedural geometry model guided by WeavingLLM, a domain-specific language model, synthesizes detailed yarn-level microstructures with natural irregularities. Our results demonstrate significantly improved detail, realism, and semantic alignment compared to state-of-the-art baselines, making it suitable for digital human clothing, interior design, and other applications. Quantitative metrics and user studies confirm the effectiveness of our approach. Extending our method to a broader range of textile types remains an open direction for future work.

\section*{Acknowledgment}
We thank the reviewers for the valuable comments. This work has been partially supported by the National Natural Science Foundation of China under grant No. 62572230.




\clearpage
\onecolumn
\twocolumn

{
    \small
    \bibliographystyle{ieeenat_fullname}
    \bibliography{main}

@String(CVPR= {IEEE Conf. Comput. Vis. Pattern Recog.})

@String(TOG= {ACM Trans. Graph.})

@String(CVPR  = {CVPR})

@String(TOG   = {ACM TOG})

@article{IrawanAndMarschner2012,
    title = {Specular Reflection from Woven Cloth},
    author = {Irawan, Piti and Marschner, Steve},
    journal = {ACM Trans. Graph.},
    volume = {31},
    number = {1},
    year = {2012},
	pages = {1-20},
	numpages = {20}
}

@article{he2024dresscode,
      title={DressCode: Autoregressively Sewing and Generating Garments from Text Guidance},
      author={He, Kai and Yao, Kaixin and Zhang, Qixuan and Yu, Jingyi and Liu, Lingjie and Xu, Lan},
      journal={ACM Transactions on Graphics (TOG)},
      volume={43},
      number={4},
      pages={1--13},
      year={2024},
      publisher={ACM New York, NY, USA}
    }

@misc{flux2024,
    author={Black Forest Labs},
    title={FLUX},
    year={2024},
    howpublished={\url{https://github.com/black-forest-labs/flux}},
}

@article{vecchio2024controlmat,
  title={Controlmat: a controlled generative approach to material capture},
  author={Vecchio, Giuseppe and Martin, Rosalie and Roullier, Arthur and Kaiser, Adrien and Rouffet, Romain and Deschaintre, Valentin and Boubekeur, Tamy},
  journal={ACM Transactions on Graphics},
  volume={43},
  number={5},
  pages={1--17},
  year={2024},
  publisher={ACM New York, NY}
}

@inproceedings{yeh2023navigating,
  title={Navigating text-to-image customization: From lycoris fine-tuning to model evaluation},
  author={Yeh, Shih-Ying and Hsieh, Yu-Guan and Gao, Zhidong and Yang, Bernard BW and Oh, Giyeong and Gong, Yanmin},
  booktitle={The Twelfth International Conference on Learning Representations},
  year={2023}
}

@inproceedings{Jin:2022:inverse,
  author = {Wenhua Jin and Beibei Wang and Milo\v{s} Ha\v{s}an and Yu Guo and Steve Marschner and Ling-Qi Yan},
  title = {Woven Fabric Capture from a Single Photo},
  booktitle={Proceedings of SIGGRAPH Asia 2022},
  year={2022}
}

@article{Wang:2023:SpongeCake,
  author = {Beibei Wang and Wenhua Jin and Milo\v{s} Ha\v{s}an and Ling-Qi Yan},
  title = {SpongeCake: A Layered Microflake Surface Appearance Model},
  journal ={ACM Transactions on Graphics},
  year = {2022},
  issue_date = {February 2023},
  volume = {42},
  number = {1},
  articleno = {8},
  numpages = {16},
}

@article{Montazeri2020ply,
author = {Montazeri, Zahra and Gammelmark, S\o{}ren B. and Zhao, Shuang and Jensen, Henrik Wann},
title = {A practical ply-based appearance model of woven fabrics},
year = {2020},
issue_date = {December 2020},
publisher = {Association for Computing Machinery},
address = {New York, NY, USA},
volume = {39},
number = {6},
issn = {0730-0301},
url = {https://doi.org/10.1145/3414685.3417777},
doi = {10.1145/3414685.3417777},
journal = {ACM Trans. Graph.},
month = nov,
articleno = {251},
numpages = {13}
}

@article{zhu2023yarn,
  title         = {A Practical and Hierarchical Yarn-based Shading Model for Cloth},
  author        = {Junqiu Zhu and Zahra Montazeri and Julien Aubry and  Ling-Qi Yan and Andrea Weidlich},
  month         = {July},
  journal       = {Computer Graphics Forum},
  year          = {2023},
  volume        = {42},
  number        = {4},
  pages         = {2--11},
  doi           = {10.1111/cgf.14894},
}

@article{Zhao2016fittingyarn,
author = {Zhao, Shuang and Luan, Fujun and Bala, Kavita},
title = {Fitting procedural yarn models for realistic cloth rendering},
year = {2016},
issue_date = {July 2016},
publisher = {Association for Computing Machinery},
address = {New York, NY, USA},
volume = {35},
number = {4},
issn = {0730-0301},
url = {https://doi.org/10.1145/2897824.2925932},
doi = {10.1145/2897824.2925932},
month = jul,
articleno = {51},
numpages = {11}
}

@misc{hesselCLIPScoreReferencefreeEvaluation2022,
  title = {{{CLIPScore}}: {{A Reference-free Evaluation Metric}} for {{Image Captioning}}},
  shorttitle = {{{CLIPScore}}},
  author = {Hessel, Jack and Holtzman, Ari and Forbes, Maxwell and Bras, Ronan Le and Choi, Yejin},
  year = {2022},
  number = {arXiv:2104.08718},
  eprint = {2104.08718},
  primaryclass = {cs},
  publisher = {arXiv},
  doi = {10.48550/arXiv.2104.08718},
  archiveprefix = {arXiv}
}

@article{ma2025materialpicker,
  title={MaterialPicker: Multi-Modal DiT-Based Material Generation},
  author={Ma, Xiaohe and Deschaintre, Valentin and Ha{\v{s}}an, Milo{\v{s}} and Luan, Fujun and Zhou, Kun and Wu, Hongzhi and Hu, Yiwei},
  journal={ACM Transactions on Graphics (TOG)},
  volume={44},
  number={4},
  pages={1--12},
  year={2025},
  publisher={ACM New York, NY, USA}
}

@inproceedings{xue2024reflectancefusion,
  title={ReflectanceFusion: Diffusion-based text to SVBRDF Generation},
  author={Xue, Bowen and Guarnera, Claudio and Zhao, Shuang and Montazeri, Zahra},
  booktitle={Eurographics Symposium on Rendering},
  year={2024},
  organization={Eurographics Association}
}

@inproceedings{GNNRaghavan,
author = {Raghavan, Nithin and Mullia, Krishna and Trevithick, Alexander and Luan, Fujun and Ha\v{s}an, Milo\v{s} and Ramamoorthi, Ravi},
title = {Generative Neural Materials},
year = {2025},
isbn = {9798400715402},
publisher = {Association for Computing Machinery},
address = {New York, NY, USA},
url = {https://doi.org/10.1145/3721238.3730746},
doi = {10.1145/3721238.3730746},
booktitle = {Proceedings of the Special Interest Group on Computer Graphics and Interactive Techniques Conference Conference Papers},
articleno = {162},
numpages = {11},
location = {
},
series = {SIGGRAPH Conference Papers '25}
}

@misc{bergmannLearningTextureManifolds2017,
  title = {Learning {{Texture Manifolds}} with the {{Periodic Spatial GAN}}},
  author = {Bergmann, Urs and Jetchev, Nikolay and Vollgraf, Roland},
  year = {2017},
  number = {arXiv:1705.06566},
  eprint = {1705.06566},
  primaryclass = {cs},
  publisher = {arXiv},
  doi = {10.48550/arXiv.1705.06566},
  archiveprefix = {arXiv}
}

@article{ying2025chord,
  title={Chord: Chain of Rendering Decomposition for PBR Material Estimation from Generated Texture Images},
  author={Ying, Zhi and Rong, Boxiang and Wang, Jingyu and Xu, Maoyuan},
  journal={arXiv preprint arXiv:2509.09952},
  year={2025}
}

@ARTICLE{Schröder2015Reverse,
  author={Schröder, Kai and Zinke, Arno and Klein, Reinhard},
  journal={IEEE Transactions on Visualization and Computer Graphics}, 
  title={Image-Based Reverse Engineering and Visual Prototyping of Woven Cloth}, 
  year={2015},
  volume={21},
  number={2},
  pages={188-200},
  doi={10.1109/TVCG.2014.2339831}
}

@misc{Li2024Fiber,
      title={Fiber-level Woven Fabric Capture from a Single Photo}, 
      author={Zixuan Li and Pengfei Shen and Hanxiao Sun and Zibo Zhang and Yu Guo and Ligang Liu and Ling-Qi Yan and Steve Marschner and Milos Hasan and Beibei Wang},
      year={2024},
      eprint={2409.06368},
      archivePrefix={arXiv},
      primaryClass={cs.GR},
      url={https://arxiv.org/abs/2409.06368}, 
}

@inproceedings{Tang2024Woven,
author = {Tang, Yingjie and Li, Zixuan and Hasan, Milos and Yang, Jian and Wang, Beibei},
title = {Woven Fabric Capture with a Reflection-Transmission Photo Pair},
year = {2024},
isbn = {9798400705250},
publisher = {Association for Computing Machinery},
address = {New York, NY, USA},
url = {https://doi.org/10.1145/3641519.3657410},
doi = {10.1145/3641519.3657410},
booktitle = {ACM SIGGRAPH 2024 Conference Papers},
articleno = {20},
numpages = {10},
location = {Denver, CO, USA},
series = {SIGGRAPH '24}
}

@inproceedings{Wu2017realtime,
author = {Wu, Kui and Yuksel, Cem},
title = {Real-time fiber-level cloth rendering},
year = {2017},
isbn = {9781450348867},
publisher = {Association for Computing Machinery},
address = {New York, NY, USA},
url = {https://doi.org/10.1145/3023368.3023372},
doi = {10.1145/3023368.3023372},
booktitle = {Proceedings of the 21st ACM SIGGRAPH Symposium on Interactive 3D Graphics and Games},
articleno = {5},
numpages = {8},
location = {San Francisco, California},
series = {I3D '17}
}

@inproceedings{sartorMatFusionGenerativeDiffusion2023a,
  title = {{{MatFusion}}: {{A Generative Diffusion Model}} for {{SVBRDF Capture}}},
  shorttitle = {{{MatFusion}}},
  booktitle = {{{SIGGRAPH Asia}} 2023 {{Conference Papers}}},
  author = {Sartor, Sam and Peers, Pieter},
  year = {2023},
  eprint = {2406.06539},
  primaryclass = {cs},
  pages = {1--10},
  doi = {10.1145/3610548.3618194},
  archiveprefix = {arXiv}
}

@article{vecchio2024stablematerials,
  title={StableMaterials: Enhancing Diversity in Material Generation via Semi-Supervised Learning},
  author={Vecchio, Giuseppe},
  journal={arXiv preprint arXiv:2406.09293},
  year={2024}
}

@article{rodriguez-pardoSeamlessGANSelfSupervisedSynthesis2023b,
  title = {{{SeamlessGAN}}: {{Self-Supervised Synthesis}} of {{Tileable Texture Maps}}},
  shorttitle = {{{SeamlessGAN}}},
  author = {{Rodriguez-Pardo}, Carlos and Garces, Elena},
  year = {2023},
  journal = {IEEE Transactions on Visualization and Computer Graphics},
  volume = {29},
  number = {6},
  eprint = {2201.05120},
  primaryclass = {cs},
  pages = {2914--2925},
  doi = {10.1109/TVCG.2022.3143615},
  archiveprefix = {arXiv}
}

@misc{zhouTileGenTileableControllable2022a,
  title = {{{TileGen}}: {{Tileable}}, {{Controllable Material Generation}} and {{Capture}}},
  shorttitle = {{{TileGen}}},
  author = {Zhou, Xilong and Ha{\v s}an, Milo{\v s} and Deschaintre, Valentin and Guerrero, Paul and Sunkavalli, Kalyan and Kalantari, Nima},
  year = {2022},
  number = {arXiv:2206.05649},
  eprint = {2206.05649},
  primaryclass = {cs},
  publisher = {arXiv},
  doi = {10.48550/arXiv.2206.05649},
  archiveprefix = {arXiv}
}

@inproceedings{zhang2024fabricdiffusion,
    title     = {{FabricDiffusion}: High-Fidelity Texture Transfer for 3D Garments Generation from In-The-Wild Images},
    author    = {Zhang, Cheng and Wang, Yuanhao and Vicente Carrasco, Francisco and Wu, Chenglei and 
                 Yang, Jinlong and Beeler, Thabo and De la Torre, Fernando},
    booktitle = {ACM SIGGRAPH Asia},
    year      = {2024},
}

@inproceedings{vecchioMatFuseControllableMaterial2024,
  title = {{{MatFuse}}: {{Controllable Material Generation}} with {{Diffusion Models}}},
  shorttitle = {{{MatFuse}}},
  booktitle = {2024 {{IEEE}}/{{CVF Conference}} on {{Computer Vision}} and {{Pattern Recognition}} ({{CVPR}})},
  author = {Vecchio, Giuseppe and Sortino, Renato and Palazzo, Simone and Spampinato, Concetto},
  year = {2024},
  eprint = {2308.11408},
  primaryclass = {cs},
  pages = {4429--4438},
  doi = {10.1109/CVPR52733.2024.00424},
  archiveprefix = {arXiv}
}

@misc{karrasAnalyzingImprovingImage2020,
  title = {Analyzing and {{Improving}} the {{Image Quality}} of {{StyleGAN}}},
  author = {Karras, Tero and Laine, Samuli and Aittala, Miika and Hellsten, Janne and Lehtinen, Jaakko and Aila, Timo},
  year = {2020},
  number = {arXiv:1912.04958},
  eprint = {1912.04958},
  primaryclass = {cs},
  publisher = {arXiv},
  doi = {10.48550/arXiv.1912.04958},
  archiveprefix = {arXiv}
}

@misc{HandWeaving.net,
  title        = {HandWeaving.net},
  author       = {Kris Bruland},
  year         = 2025,
  note         = {\url{https://handweaving.net/}}
}

@misc{qwen2.5,
    title = {Qwen2.5: A Party of Foundation Models},
    url = {https://qwenlm.github.io/blog/qwen2.5/},
    author = {Qwen Team},
    month = {September},
    year = {2024}
}

@misc{dettmers2023qlora,
      title={QLoRA: Efficient Finetuning of Quantized LLMs}, 
      author={Tim Dettmers and Artidoro Pagnoni and Ari Holtzman and Luke Zettlemoyer},
      year={2023},
      eprint={2305.14314},
      archivePrefix={arXiv},
      primaryClass={cs.LG},
      url={https://arxiv.org/abs/2305.14314}, 
}

@misc{openai_gpt5_2025,
  title        = {Introducing {GPT-5}},
  author       = {{OpenAI}},
  year         = {2025},
  howpublished = {\url{https://openai.com/index/introducing-gpt-5/}},
  note         = {Accessed: YYYY-MM-DD}
}

@inproceedings{zhang2018lpips,
  title={The Unreasonable Effectiveness of Deep Features as a Perceptual Metric},
  author={Zhang, Richard and Isola, Phillip and Efros, Alexei A and Shechtman, Eli and Wang, Oliver},
  booktitle={CVPR},
  year={2018}
}

@inproceedings{Radford2021LearningTV,
  title={Learning Transferable Visual Models From Natural Language Supervision},
  author={Alec Radford and Jong Wook Kim and Chris Hallacy and Aditya Ramesh and Gabriel Goh and Sandhini Agarwal and Girish Sastry and Amanda Askell and Pamela Mishkin and Jack Clark and Gretchen Krueger and Ilya Sutskever},
  booktitle={International Conference on Machine Learning},
  year={2021},
  url={https://api.semanticscholar.org/CorpusID:231591445}
}
}


\end{document}